\documentclass[runningheads]{llncs}
\pdfoutput=1
\usepackage{graphicx}
\usepackage{subfigure}
\usepackage{amsmath}

\begin{document}

\title{Towards Interpretable Deep Learning Models for Knowledge Tracing}
%
\author{Yu Lu\inst{1,2}\and
	Deliang Wang\inst{2}\and
	Qinggang Meng\inst{1}\and
	Penghe Chen\inst{1}
}
\authorrunning{Yu Lu et al.}
\institute{Advanced Innovation Center for Future Education \and
	School of Educational Technology, Beijing Normal University, China
	\email{\{luyu,chenpenghe\}@bnu.edu.cn}\\
}
\maketitle              
\begin{abstract}
As an important technique for modeling the knowledge states of learners, the traditional knowledge tracing (KT) models have been widely used to support intelligent tutoring systems and MOOC platforms. Driven by the fast advancements of deep learning techniques, deep neural network has been recently adopted to design new KT models for achieving better prediction performance. However, the lack of interpretability of these models has painfully impeded their practical applications, as their outputs and working mechanisms suffer from the intransparent decision process and complex inner structures. We thus propose to adopt the post-hoc method to tackle the interpretability issue for deep learning based knowledge tracing (DLKT) models. Specifically, we focus on applying the layer-wise relevance propagation (LRP) method to interpret RNN-based DLKT model by backpropagating the relevance from the model's output layer to its input layer. The experiment results show the feasibility using the LRP method for interpreting the DLKT model's predictions, and partially validate the computed relevance scores from both question level and concept level. We believe it can be a solid step towards fully interpreting the DLKT models and promote their practical applications in the education domain.

\end{abstract}
\section{Introduction}
Knowledge tracing (KT) is a machine learning technique that attempts to model the knowledge states of learners by quantitatively diagnosing their mastery level on individual concept (e.g., ``integer” or ``fraction” in algebra), where learner's past exercise data are utilized to train the KT models. The traditional KT models, such as Bayesian knowledge tracing (BKT)~\cite{corbett1994knowledge}, treat knowledge state as a latent variable, and estimate learner's mastery probability utilizing the observed exercise results from learners. Given a learner’s consecutive exercise results (either correct or incorrect) on a sequence of questions, the KT models could timely update the learner's mastery probability on the concept level, and subsequently make prediction on the learner’s future performance. The estimated knowledge state can be directly used to evaluate a learner's current strength and weakness, and thus the KT models serve as a key component of today's intelligent tutoring system (ITS)~\cite{yazdani1989air} and massive open online course (MOOC) platforms~\cite{pardos2013adapting,wang2016structured}, which mainly provide personalized scaffolding and online learning supports. 


The rapid development of ITS and MOOC platforms greatly facilitates building KT models by collecting a large size of learner's learning and exercise data in a rapid and inexpensive way. Yet, the collected massive and consecutive exercise questions are usually associated with multiple concepts, and the traditional KT models cannot well handle the questions without explicit labels and capture the relationships among a large size of concepts (e.g., 100 or more concepts). Accordingly, deep learning models are recently introduced into the KT domain because of their powerful representation capability~\cite{piech2015deep}. Given the sequential and temporal characteristics of learner's exercise data, the recurrent neural network (RNN)~\cite{schuster1997bidirectional} is frequently adopted for building the deep learning based knowledge tracing (DLKT) models. Since it is difficult to directly measure the actual knowledge state of a learner, the existing DLKT models often adopt an alternative solution that minimizes the difference between the predicted and the real responses on exercise questions. Hence, the major output of DLKT models are the predicted performance on next questions. As a popular implementation variants of RNN, the long short-term memory (LSTM) unit~\cite{hochreiter1997long} and GRU~\cite{cho2014properties} are widely used in the DLKT models, and have achieved comparable or even better prediction performance in comparison to the traditional KT models~\cite{piech2015deep,chen2018prerequisite}.

Similar as the deep learning models operating as a ``black-box" in many other domains~\cite{monta18DSP}, the existing DLKT models also suffer from the interpretability issue, which has painfully impeded the practical applications of DLKT models in the education domain. The main reason is that it is principally hard to map a deep learning model's abstract decision (e.g. predicting correct on next question) into the target domain that end-users could easily make sense of (e.g., enabling the ITS designers or users to understand why predicting correct on next question). In this work, we attempt to tackle the above issue by introducing the proper interpreting method for the DLKT models. In particular, we adopt a post-hoc interpreting method as the tool to understand and explain the RNN-based DLKT models, where the layer-wise relevance propagation (LRP) method is utilized to backpropagate the relevance from the model's output layer to its input layer. We conduct the extensive experiments to understand the interpreting results from both question level and concept level. The experiment results validate the feasibility of interpreting the DLKT model's predictions and visualize the captured relationships among the concepts associated with the questions. By demonstrating the promise of applying the proper interpreting methods into the KT domain, this work can be a solid step towards building the fully interpretable and explainable DLKT models for the education domain. 


\section{Related Work}\label{sec:related-work}

\subsection{Knowledge Tracing with Deep Learning Models}
Early studies adopt different machine learning models to conduct KT, where BKT can be regarded as the most prominent one. The BKT model mainly adopts hidden Markov model (HMM) to estimate learner's mastery state on individual concept, and the subsequent studies improve the model by considering cognitive factors~\cite{d2008more}, learner's ability~\cite{yudelson2013individualized,liu2017towards}, knowledge prior~\cite{chen2017tracking}, difficulty level of questions~\cite{pardos2011kt} and learning time~\cite{baker2011detecting}. Besides BTK model, factor analysis~\cite{pavlik2009performance,cen2006learning} and matrix factorization-based~\cite{vie2019knowledge,thai2012factorization} models have been proposed and investigated as well. In addition,  researchers recently also adopt functional magnetic resonance imaging (fMRI) signals to improve knowledge tracing models~\cite{Halpern18EDM}.

As indicated earlier, deep learning models are recently introduced into the KT domain, as they have enough capacity to automatically learn the inherent relationships and do not require explicit labels on the concept level. Deep knowledge tracing (DKT)~\cite{piech2015deep} that utilizes LSTM can be regarded as the pioneer work, while some limitations have been reported~\cite{xiong2016going}. Subsequently, The dynamic key-value memory network~\cite{zhang2017dynamic} and its variants~\cite{chaudhry2018modeling,yeung2019edm} are adopted to improve KT performance, which are mainly based on memory-augmented neural networks~\cite{santoro2016icml}. Meanwhile, the attention network~\cite{su2018exercise} has been also introduced to better represent question semantics. Besides, prerequisite~\cite{chen2018prerequisite}, prediction consistent regularization~\cite{yeung2018addressing} and image~\cite{liu2018kdd} information have been considered and utilized to design DLKT models as well. In short, deep learning models has been adopted and played a crucial role for addressing KT problem.

\subsection{Interpreting Deep Learning Models}
Although deep learning models have achieved outstanding performance in KT, the lack of interpretability has painfully impeded the practical applications of these models. To tackle such a critical issue, researchers have proposed a number of methods to help end-users understand the model outputs and even its inner working mechanism. The interpretability can be categorized into \textit{ante-hoc} and \textit{post-hoc} interpretabilities: the \textit{ante-hoc} interpretability refers to training self-explainable machine learning models~\cite{melis2018towards}, whereas it is usually constrained by using simple-structured models, such as linear regression~\cite{strumbelj2010an}, decision tree~\cite{deng2019IJDSA} or naive Bayes models~\cite{poulin2006visual} and hard to handle the complex deep learning models. 


Among different types of \textit{post-hoc} interpretability, the \textit{local} methods mainly focus on understanding the model's decisions or predictions from the perspective of its readily input variables. Specifically, given a model's decision (e.g., classification or regression results), the \textit{local} methods usually analyze the contributions of input variable's features to explain that decision. The typical \textit{local} methods include backward propagation~\cite{zeiler2014eccv}, sensitivity analysis~\cite{Saltelli04Wiley}, simple Taylor decomposition~\cite{lapuschkin2016cvpr} and feature inversion methods~\cite{du2018towards}. For example, backward propagation method interprets a model's decision by explicitly using its deep neural network structure and the backpropagation mechanism~\cite{rumelhart1988learning}, which usually starts from the model output, progressively and inversely maps the prediction onto the lower layers until it reaches the model input. The LRP method~\cite{bach2015po} can be regarded as a typical backward propagation method, where the share of model output received by each neuron is properly redistributed by its predecessors to achieve the relevance conservation, and the injection of negative relevance is controlled by its hyperparameters. LRP method is applicable and empirically scales to general deep learning models. It has been adopted for image classification~\cite{arbabzadah2016cv}, machine translation~\cite{ding2017acl} and text analysis~\cite{arras2017ps}.

In short, the research on interpreting machine learning models, especially on the complex and domain-specific deep learning models, is still in its infancy. In the education domain, researchers have started interpreting KT models~\cite{yang2018implicit}, but most studies target on the traditional simple-structured Bayesian
network-based ones~\cite{baker08its,qiu2011edm}. In this work, we mainly focus on explaining the DLKT models by leveraging on \textit{local} interpretability method.

\section{Interpreting RNN-based KT Model}\label{sec:interprete-model}
As mentioned earlier, KT models need to handle the sequential and temporal characteristics of learner's exercise data, and thus the RNN is often adopted, especially for the DLKT models. In this section, we first briefly introduce the RNN-based KT models, and then present an effectively local interpretability method, namely LRP method, and finally take the LRP to conduct the interpreting task.  

\subsection{RNN-based DLKT Model}
\begin{figure}[!t]
	\centering
	\includegraphics[width=0.95\textwidth]{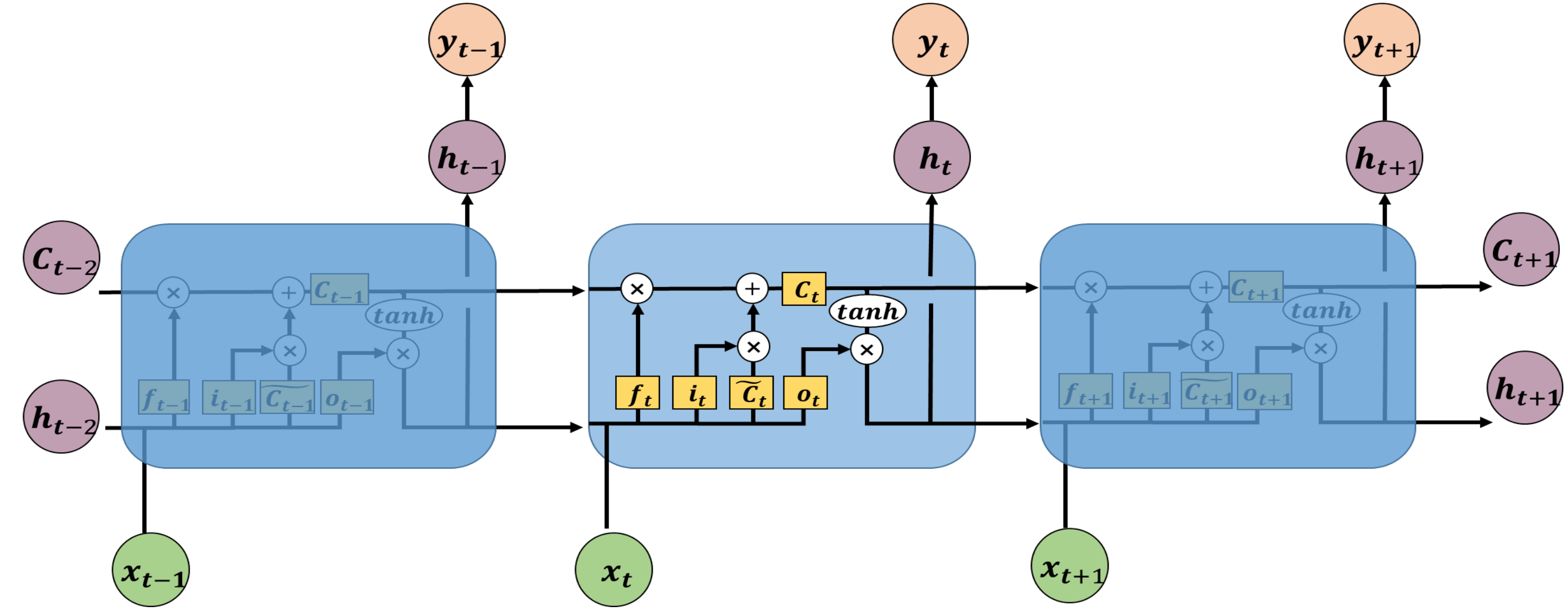}
	\caption{The architecture of a LSTM-based DLKT model} 
	\label{fig:DKT-Arch}
\end{figure}

Briefly speaking, RNN is a practical and effective implementation of neuron architecture that is able to connect previous information to the present task, such as linking the learner's past exercise information to her performance on next questions. Hence, a number of DLKT models, such as DKT~\cite{piech2015deep}, adopt LSTM or similar architectures (e.g., GRU) to accomplish the KT task, as illustrated in Figure~\ref{fig:DKT-Arch}. As a typical RNN architecture, the model maps an input sequence vectors~\{$\bf x_0, ..., x_{t-1}, x_t,...$\} to an output sequence vectors~\{$\bf y_0,..., y_{t-1}, y_t,...$\}, where~$\bf x_t$ represents the interaction between learners and exercises, and~$\bf y_t$ refers to the predicted probability vectors on mastering the concepts. The standard LSTM unit is usually implemented in the DLKT models as follows:

\begin{eqnarray}\label{eqn:ft}
\vspace{-3mm}
{{\rm{f}}_{\rm{t}}} = {\rm{\sigma }}\left( {{{\bf{W}}_{{\rm{fh}}}}{{\rm{h}}_{{\rm{t}} - 1}} + {{\bf{W}}_{{\rm{fx}}}}{{\bf{x}}_{\rm{t}}} + {{\rm{b}}_{\rm{f}}}} \right)\\
{{\rm{i}}_{\rm{t}}} = {\rm{\sigma }}\left( {{{\bf{W}}_{{\rm{ih}}}}{{\rm{h}}_{{\rm{t}} - 1}} + {{\bf{W}}_{{\rm{ix}}}}{{\bf{x}}_{\rm{t}}} + {{\rm{b}}_{\rm{i}}}} \right)\\
\widetilde {{{\rm{C}}_{\rm{t}}}} = {\rm{tanh}}\left( {{{\bf{W}}_{{\rm{ch}}}}{{\rm{h}}_{{\rm{t}} - 1}} + {{\bf{W}}_{{\rm{cx}}}}{{\bf{x}}_{\rm{t}}} + {{\rm{b}}_{\rm{c}}}} \right)\\
{{\rm{C}}_{\rm{t}}} = {{\rm{f}}_{\rm{t}}}\odot{{\rm{C}}_{{\rm{t}} - 1}} + {{\rm{i}}_{\rm{t}}}\odot\widetilde {{{\rm{C}}_{\rm{t}}}}\\
{{\rm{o}}_{\rm{t}}} = {\rm{\sigma }}\left( {{{\bf{W}}_{{\rm{oh}}}}{{\rm{h}}_{{\rm{t}} - 1}} + {{\bf{W}}_{{\rm{ox}}}}{{\bf{x}}_{\rm{t}}} + {{\rm{b}}_{\rm{o}}}} \right)\\
{{\rm{h}}_{\rm{t}}} = {{\rm{o}}_{\rm{t}}}\odot{\rm{tanh}}\left( {{{\rm{C}}_{\rm{t}}}} \right).
\vspace{-3mm}
\end{eqnarray}

After getting the LSTM output~$h_t$, the DLKT models usually further adopt an additional layer to output the final predicted results~$y_t$ as below:

\begin{equation}
{{\bf{y}}_{\rm{t}}} = {\rm{\sigma }}\left( {{{\bf{W}}_{{\rm{yh}}}}{{\rm{h}}_{\rm{t}}} + {{\rm{b}}_{\rm{y}}}} \right).
\end{equation}

From the above implementations, we see that that the RNN-based DLKT models usually consist of two types of connections: \textbf{\textit{weighted linear connection}}, i.e., Eqn.~(1),~(2),~(3),~(5),~(7), and \textbf{\textit{multiplicative connection}}, i.e., Eqn.~(4)~and~(6). The two types would be handled by the interpreting method in a different way.

\subsection{LRP Method}
As mentioned earlier, we mainly adopt the LRP method to address the interpreting task, which mainly focuses on analyzing the contributions of input's individual features for explaining the decision. Given a trained DLKT model, the share of the model's current output received by each neuron is properly redistributed by its predecessors to achieve the relevance conservation, and the quantity being back propagated can be filtered to only retain what passes through certain neurons. Specifically, it firstly sets the relevance of the output layer neuron to the model's output value for the target class, and meanwhile simply ignores the other output layer's values and neurons. After that, it starts backpropagating the relevance score from the output layer to the input layer, where the two types of connections, i.e., weighted linear connection and multiplicative connection, in the intermediate layers would be handled in a different way.

\subsubsection{Weighted Linear Connection:} as shown in Eqn.~(1),~(2),~(3),~(5),~(7), this type of connection can be written in a general form:

\begin{equation}
a = activation\left( {\bf{W}\rm{h} + \bf{W}\bf{x} + \rm{b}} \right),
\end{equation}
where~$activation(*)$ is an activation function commonly used in the deep learning models. Assuming activation functions do not change the relevance distribution, the weighted connection can be further denoted as:

\begin{equation}
a_j^{\left( {l + 1} \right)} = \mathop \sum \limits_i {w_{ij}}{\rm{\;}}a_i^{\left( l \right)} + {b_j},
\end{equation}
where~$ a_j^{\left( {l + 1} \right)} $ is the message received in the feedforward direction at neuron $j$ in layer~$l+1$, $w_{ij}$ and~$b_j$ are the corresponding connection weight and bias. Given~$R_{i \leftarrow j}^{\left( {l} \right)}$ is the relevance received by neuron~$i$ in layer~$l$ from neuron~$j$ in layer~$l+1$, we have

\begin{equation}
R_{i \leftarrow j}^{\left( {l} \right)} = \frac{{{w_{ij}}a_i^{\left( l \right)} + \frac{{sign\left( {a_j^{\left( {l + 1} \right)}} \right)\;\varepsilon \; + {b_j}\;}}{N}\delta}}{{a_j^{\left( {l + 1} \right)}\; + \;sign\left( {a_j^{\left( {l + 1} \right)}} \right)\;\varepsilon }}\;R_j^{\left( {l + 1} \right)},
\end{equation}
where $N$ is the number of neurons in layer~$l$, and the item~$sign\left( {a_j^{\left( {l + 1} \right)}} \right)*\varepsilon$ is a stabilizer to prevent~$R_{i \leftarrow j}^{\left( {l} \right)}$ becoming an unbounded value. In practice, $\varepsilon$ can be a small positive value, and~$sign\left( {a_j^{\left( {l + 1} \right)}} \right)$ is set to 1 or -1, given ${a_j^{\left( {l + 1} \right)}}$ is positive or negative respectively. $\delta$ can be set to either 1 or 0, depending on whether it is allowed the bias~$b_j$ and stabilizer to receive the share of total relevance. In this work, we set~$\delta =0$ to conserve more relevance for the lower-level neurons. $R_{j}^{\left( {l+1} \right)}$ is the total relevance received by neuron~$j$ in the upper layer layer~$l+1$. Normally, neuron~$i$ is connected to multiple neurons in upper layer (not only neuron~$j$), and thus its total relevance can be computed as:

\begin{equation}
R_i^{\left( l \right)} = \sum\nolimits_j^{} {R_{i \leftarrow j}^{\left( l \right)}},
\end{equation}
where all the neurons~$j$ are located in upper layer~$l+1$.

\subsubsection{Multiplicative Connection:} the LTSM unit(or GRU) typically has the ``gate" structures in the form of multiplicative connection as shown in Eqn.~(4)~and~(6), where the neuron ouput ranging between 0 to 1 can be called ``gate" neuron, and the remaining one can be regarded as the ``source" neuron. This type of connection can be written in a general form:

\begin{equation}
a_j^{\left( {l + 1} \right)} = a_g^{\left( l \right)}{\; \rm{\odot}\; }a_c^{\left( l \right)},
\end{equation}
where~$a_g^{\left( l \right)}$ and $a_c^{\left( l \right)}$ are the messages received in the feedforward direction by the ``gate" neuron~$g$ and ``source" neuron~$c$ in layer~$l$. In the feedforward direction, the ``gate" neuron has decided how much of the information should be retained to in the upper-layer neurons and eventually contributed to the model's decision~\cite{arras2017explaining}, we can simply regard its relevance received from the upper layer as zero and meanwhile give the full credit to the ``source" gate. Hence, we have

\vspace{-4mm}
\begin{eqnarray}
\vspace{-4mm}
R_{c \leftarrow j}^{\left( {l} \right)} = R_{j}^{\left( {l+1} \right)};\\
R_{g \leftarrow j}^{\left( {l} \right)} = 0,
\vspace{-4mm}
\end{eqnarray}
where~$R_{c \leftarrow j}^{\left( {l} \right)}$ and~$R_{g \leftarrow j}^{\left( {l} \right)}$ are the relevance received by neuron~$c$ and neuron~$g$ in layer~$l$ from neuron~$j$ in layer~$l+1$, respectively. In short, using the above described methods to compute the back-propagated relevances for the two types of connections, we could conduct interpreting the RNN-based DLKT models. 

\begin{figure}[!t]
	\centering
	\includegraphics[width=0.95\textwidth]{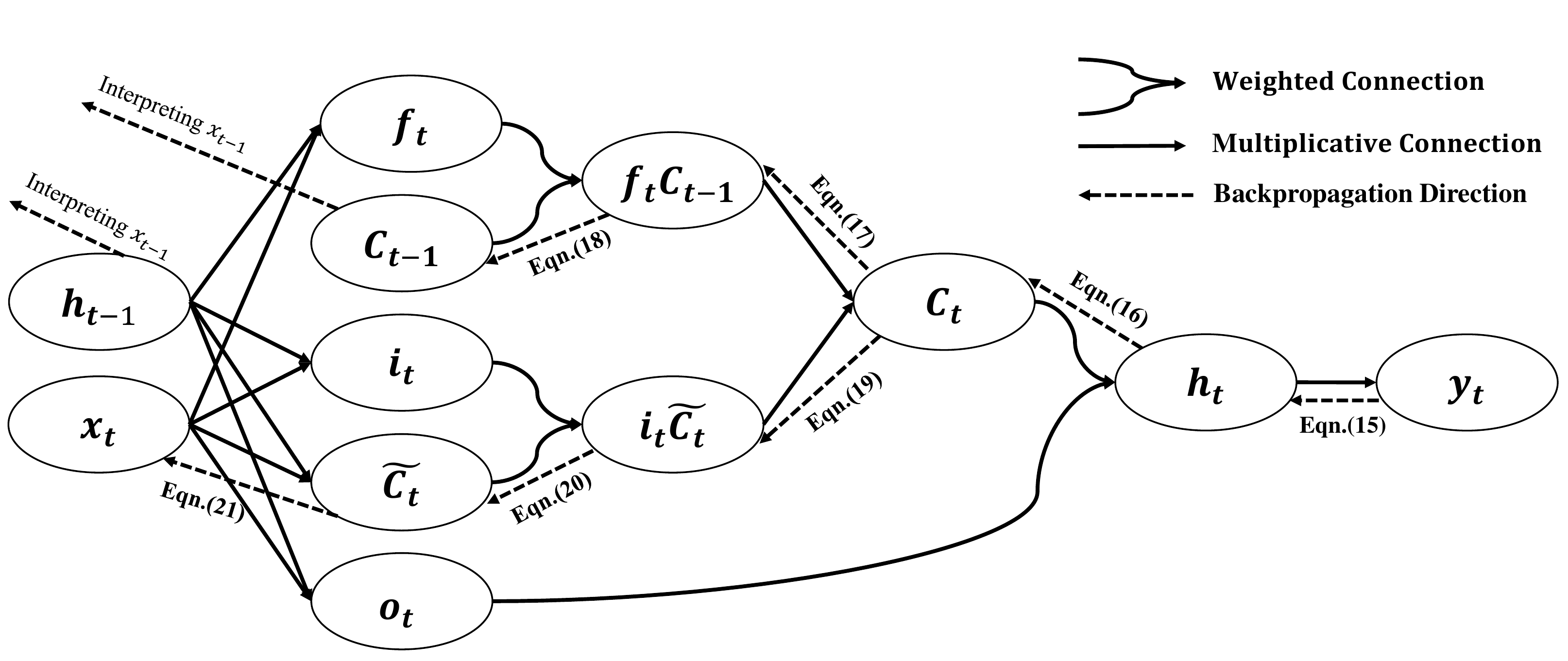}
	\vspace{-4mm}
	\caption{The feedforward prediction path with the two types of connections in LSTM and the backpropagation path for interpreting its prediction results} 
	\vspace{-4mm}
	\label{fig:Interpre-Flow}
\end{figure}

\vspace{-4mm}
\subsection{Interpreting DLKT Models using LRP Method}
Considering the RNN-based DLKT model given in Eqn.(1) to (10) and the interpreting method given in Eqn.~(10) to (11) and (13) to (14), the interpreting task can be accomplished by computing the relevance as below:

\begin{align}
& {R_{{{\rm{h}}_{\rm{t}}}}} = \frac{{\bf{W_{yh}}\rm{h_t}}}{{\bf{W_{yh}}\rm{h_t} + {b_y} + \varepsilon  * sign(\bf{W_{yh}}\rm{h_t} + {b_y})}} * R_{{y_t}}^d &\\
& {R_{{C_t}}} = {R_{{h_t}}} &\\
& {R_{{f_t}{C_{t - 1}}}} = \frac{{{f_t}{C_{t - 1}}}}{{{C_t} + \varepsilon  * sign({C_t})}} * {R_{{C_t}}} &\\
& {R_{{C_{t - 1}}}} = {R_{{f_t}{C_{t - 1}}}} &\\
& {R_{{i_t}{{\tilde C}_t}}} = \frac{{{i_t}{{\tilde C}_t}}}{{{C_t} + \varepsilon  * sign({C_t})}} * {R_{{C_t}}} &\\
& {R_{{{\tilde C}_t}}} = {R_{{i_t}{{\tilde C}_t}}} &
\end{align}
where~$R_{{y_t}}^d$ is the value of the $d^{th}$ dimension of the prediction output~$y_t$. Finally, the calculated relevance value~$R_{{x_t}}$ for the input~$x_t$ can be derived as

\begin{align}
\vspace{-4mm}
&{R_{{x_t}}} = \frac{{\bf{W_{cx}}{x_t}}}{{\bf{W_{ch}}\rm{h_{t - 1}} + \bf{W_{cx}}\bf{x_t} + \rm{b_c} + \varepsilon  * sign(\bf{W_{ch}}\rm{h_{t - 1}} + \bf{W_{cx}}{x_t} + \rm{b_c})}}*{R_{{{\tilde C}_t}}}&
\vspace{-4mm}
\end{align}

Figure~\ref{fig:Interpre-Flow} further illustrates both of the feedforward prediction path with the two types of connections, and the backpropagation direction for interpreting the prediction results. Note that the above process is generally applicable to computing the relevance of the model inputs (e.g., $x_{t-1}$), while the way of computing $R_{C_{t-1}}$ might be slightly different.

We further take an exemplary but real case to demonstrate the interpreting results. Given a trained DLKT model for math and a learner's exercise sequence as the input, the input consists of seven consecutive questions, where the questions are associated with three different math concepts respectively. Table~\ref{tab:exemplary-case} shows the question details and whether the learner correctly answers the questions. Assuming the next question is on the concept \textit{subtraction numbers}, we then get the prediction result 0.746 from the output~$y_t$, i.e., the probability of correctly answering the next question. By iteratively using the proposed interpreting method, we can finally obtain the relevance value for each input, i.e., from the 1st question to the 7th question, as shown in the last row of Table~\ref{tab:exemplary-case}. We see clearly that a significant positive relevance is obtained by the correctly-answered questions on the same or closely-related concepts (i.e., the 1st, 2nd, 6th and 7th questions), whereas a significant negative relevance is obtained by a falsely-answered question on the same concept (i.e., the 5th question). Furthermore, a small relevance is obtained by the questions on \textit{Area Rectangle} (i.e., the 3rd and 4th questions), which is obviously a geometry concept far from the target algebra concept, and thus has a limited contribution to the current prediction task. From the this exemplary case, we see a meaningful interpreting results by leveraging on the proposed method, and we then conduct the evaluations.

\section{Evaluation}\label{sec:evaluation}
\subsection{Data and DLKT Model Training}
We choose the public educational dataset ASSISTment 2009-2010\footnote{https://sites.google.com/site/assistmentsdata/home/assistment-2009-2010-data}\cite{feng2009addressing}, which has been commonly used for building the KT models. Specifically, we employ its ``skill builder" dataset for math, and filter out all the duplicate exercise sequence and those without labeling concept. Eventually, the dataset used for training the DLKT model consists of 325,637 answering records on 26,688 questions associated with 110 concepts from 4,151 students.

\begin{table}[!t]
	\centering
	\caption{An Exemplary Case Illustrating the Interpreting Results}
	\includegraphics[width=1.0\textwidth]{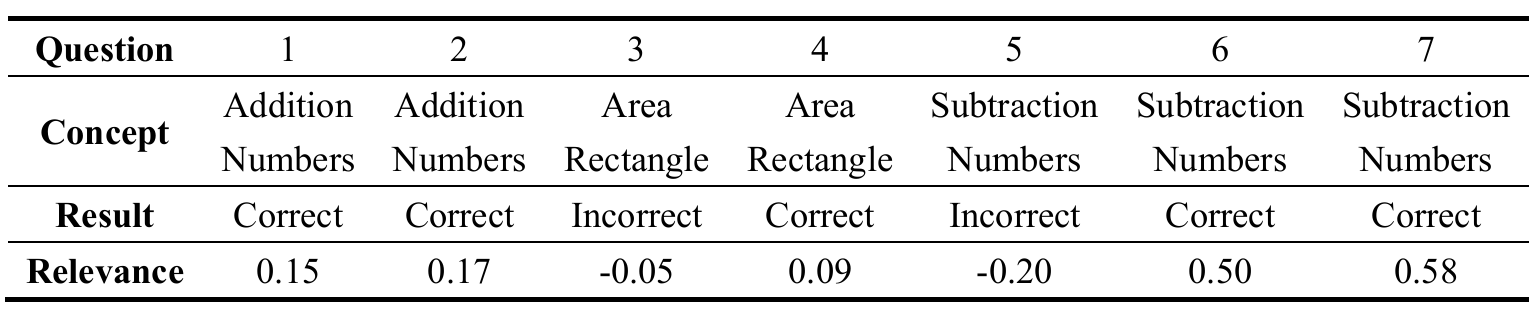}
	\label{tab:exemplary-case}
\end{table}

As shown in Figure~\ref{fig:DKT-Arch}, the built DLKT model adopts the LSTM unit with the hidden dimensionality of 256. During the training process, the mini-batch size and the dropout are set to 20 and 0.5 respectively. Adam optimization algorithm
is utilized for the model training, whose iteration number and initial learning rate are set to 500 and 0.01. Besides, 80\% data is randomly selected as the training data, and the remaining 20\% is used as the testing data. We repeat the experiment 10 times to compute the performance metrics. Considering KT as a classification problem and the exercise results as binary variables, namely 1 representing correct and 0 representing incorrect answers, the overall prediction accuracy (ACC) and AUC achieve 0.75 and 0.70 respectively. 


\subsection{Question-Level Validation}
\begin{figure}[!t]
	\centering
	\subfigure[Positive Prediction Group]{\includegraphics [width=.45\textwidth]{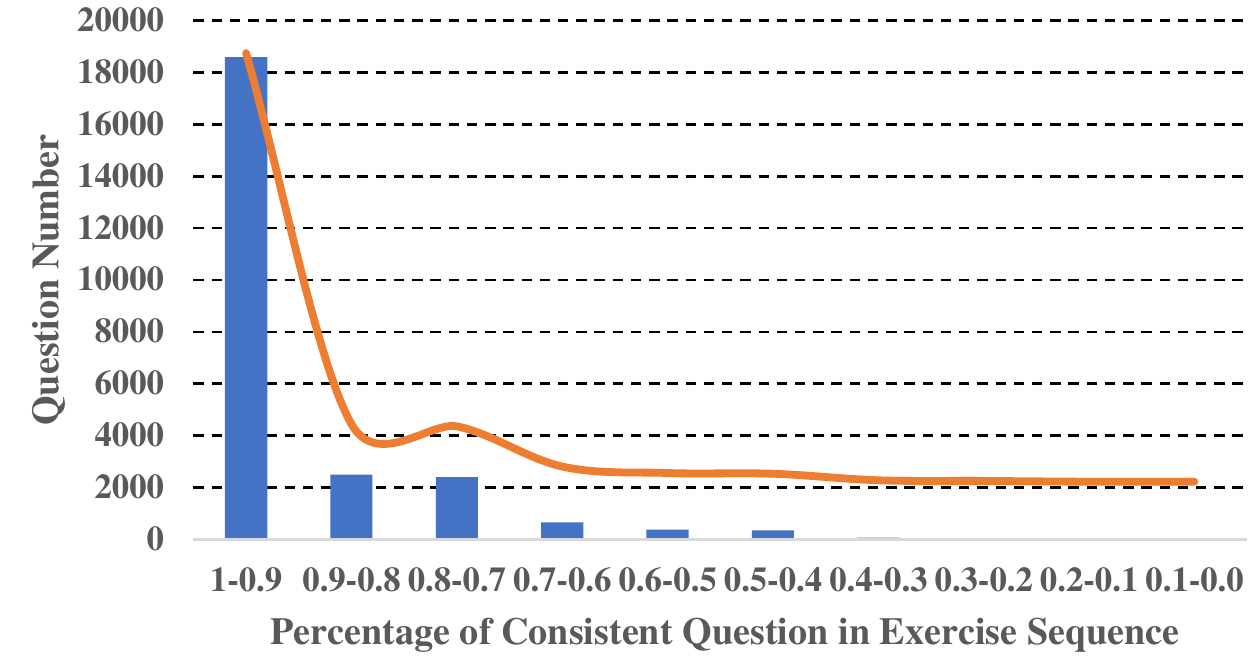} \label{sta1}}~~~~
	\subfigure[Negative Prediction Group]{\includegraphics [width=.45\textwidth]{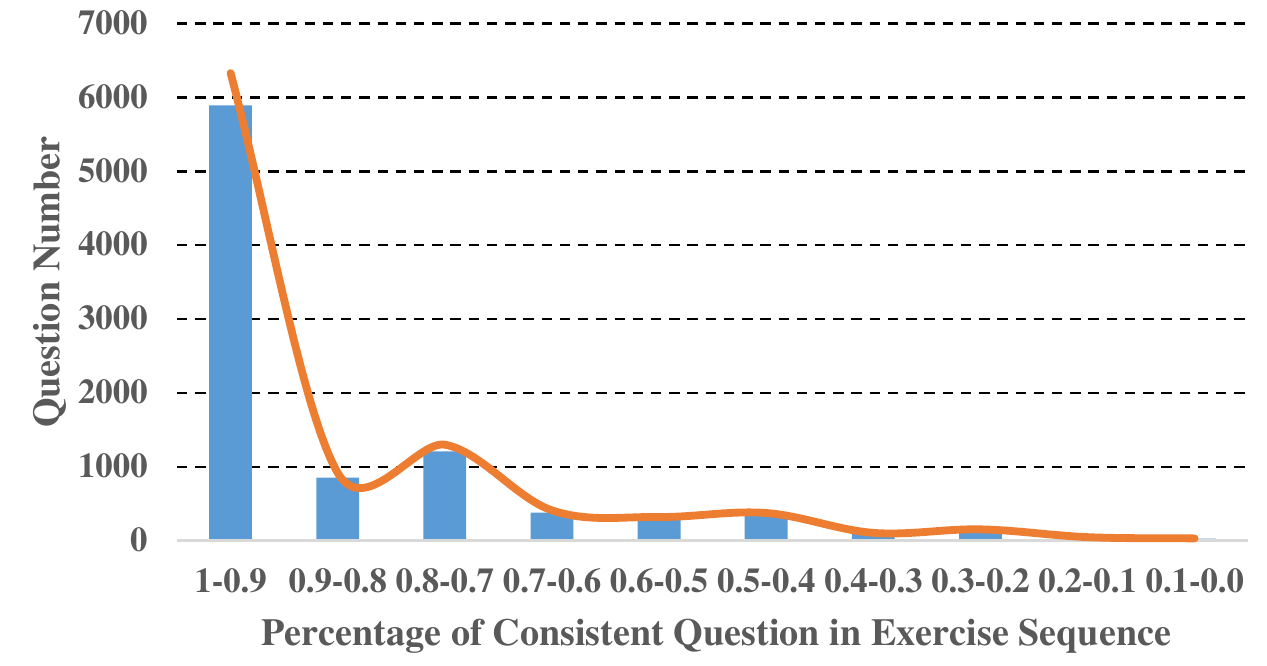} \label{sta2}}
	\caption{Histogram of the consistent rate on both positive and negative prediction groups}
	\label{fig:consistency}
\end{figure}

We firstly conduct the experiment to understand the relationship between the LRP interpreting results and the model prediction results. Specifically, we choose 48,673 exercise sequences with a length of 15, i.e., each sequence consisting of 15 individual questions, as the test dataset for the interpreting tasks. For each sequence, we take its first 14 questions as the input to the built DLKT model, and the last one to validate the model's prediction on the 15th question. As the result, the DKLT model correctly predicts the last question for 34,311 sequences, where the positive and negative results are 25,005 and 9,306 respectively. Based on the correctly predicted sequences, we adopt the LRP method to calculate the relevance values of the first 14 questions, and then investigate whether the sign of relevance values is consistent with the correctness of learner's answer. Specifically, we define \textit{consistent question} among the previous exercise questions as ``either the correctly-answered questions with a positive relevance value" or ``the falsely-answered questions with a negative relevance value". Accordingly, we compute the percentage of such consistent questions in each sequence, and name it as \textit{consistent rate}. Intuitively, a high \textit{consistent rate} reflects that most correctly-answered questions have a positive contribution and most falsely-answered questions have a negative contribution to the predicted mastery probability on the given concept. Figure~\ref{fig:consistency} shows the histogram of the consistent rate on both groups of positive prediction (i.e., the mastery probability above 50\%) and negative prediction (i.e., the mastery probability below 50\%). Clearly, we see that the majority of the exercise sequences achieve 90 percent (or above) consistent rate, which partially validates the question-level feasibility of using LRP method to interpret DLKT model's prediction results.

\begin{figure}[!t]
	\centering
	\subfigure[Positive Prediction Group]{\includegraphics [width=.48\textwidth]{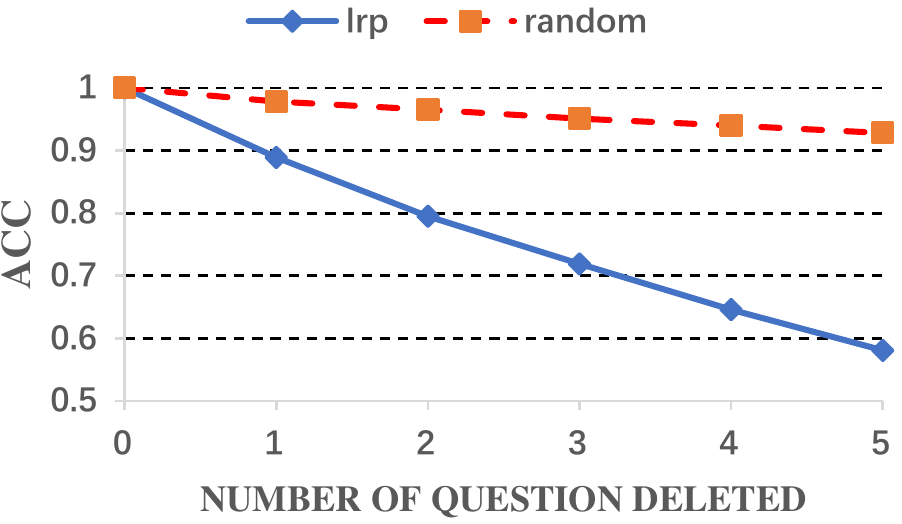} \label{del11}}~~~~
	\subfigure[Negative Prediction Group]{\includegraphics [width=.48\textwidth]{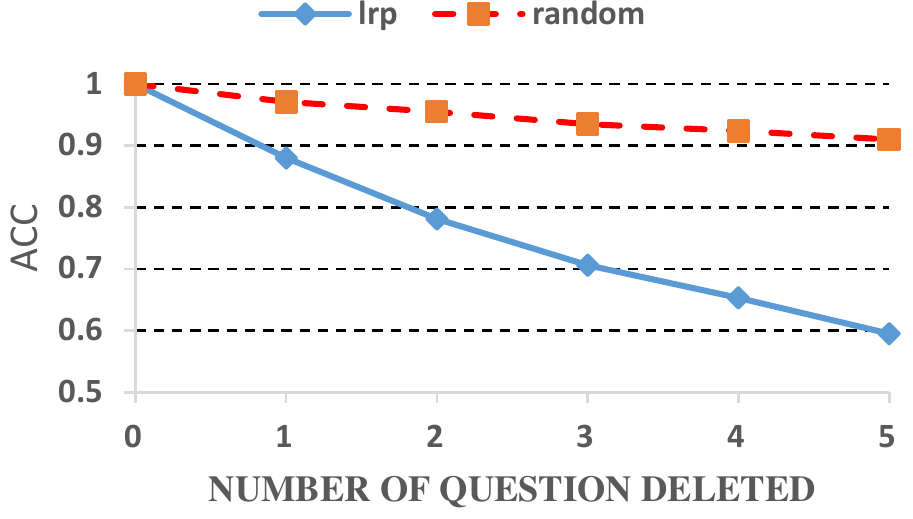} \label{del12}}
	\caption{Question deletion impact on prediction accuracy for correctly predicted ones}
	\label{fig:delet1}
\end{figure}
\begin{figure}[!t]
	\centering
	\subfigure[Positive Prediction Group]{\includegraphics [width=.48\textwidth]{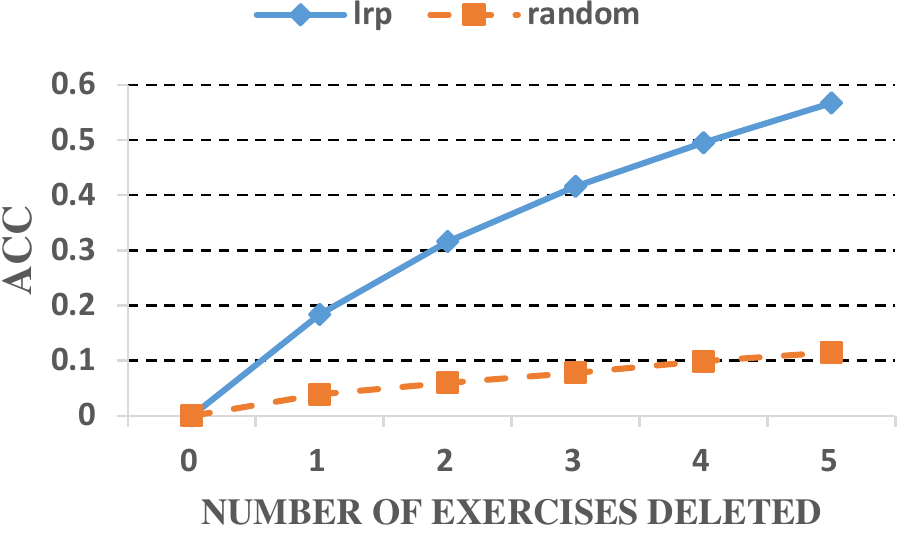} \label{del21}}~~~~
	\subfigure[Negative Prediction Group]{\includegraphics [width=.48\textwidth]{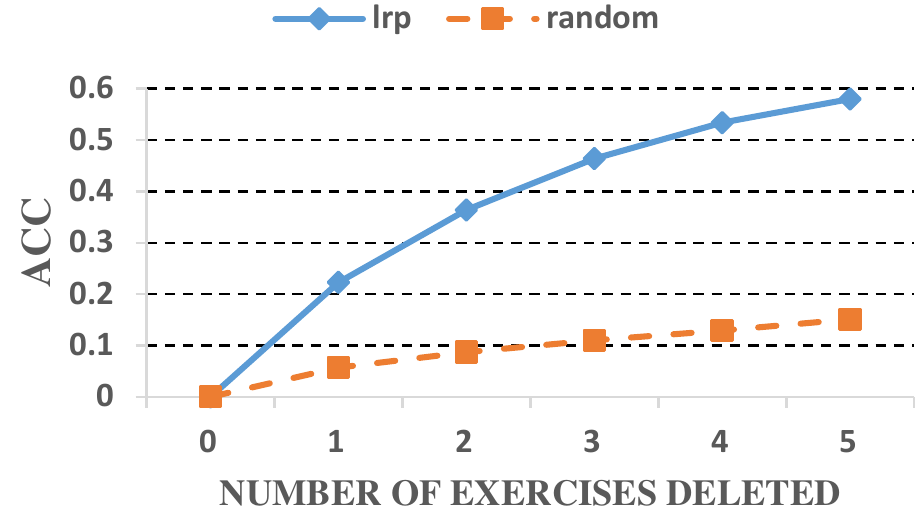} \label{del22}}
	\caption{Question deletion impact on prediction accuracy for falsely predicted ones}
	\label{fig:delet2}
\end{figure}

To further quantitatively validate the question-level relevances obtained by the LRP method, we perform the question deleting experiments. Among the correctly predicted exercise sequences, we delete questions in the decreasing order (for the positive prediction group) and increasing order (for the negative prediction group) of their relevance values, respectively. Among the falsely predicted exercise sequences, we delete questions in the decreasing order (for the positive prediction group) and increasing order (for the negative prediction group) of their relevance values, respectively. Meanwhile, we additionally perform a random question deletion for the comparison purpose for all the experiments. Figure~\ref{fig:delet1} and~\ref{fig:delet2} illustrate the results in terms of tracking the prediction accuracy over the number of question deletions. We see that in Figure~\ref{fig:delet1}, deleting questions significantly decreases the DLKT model's performance on those correctly predicted sequences, and meanwhile in Figure~\ref{fig:delet2}, deleting questions significantly improves the DLKT model's performance on those falsely predicted sequences. The question deleting results partially validate the LRP method could properly compute the question-level contributions for interpreting the DLKT model's predictions.

\subsection{Concept-Level Relationship}

\begin{figure}[!t]
	\centering
	\includegraphics[width=0.99\textwidth]{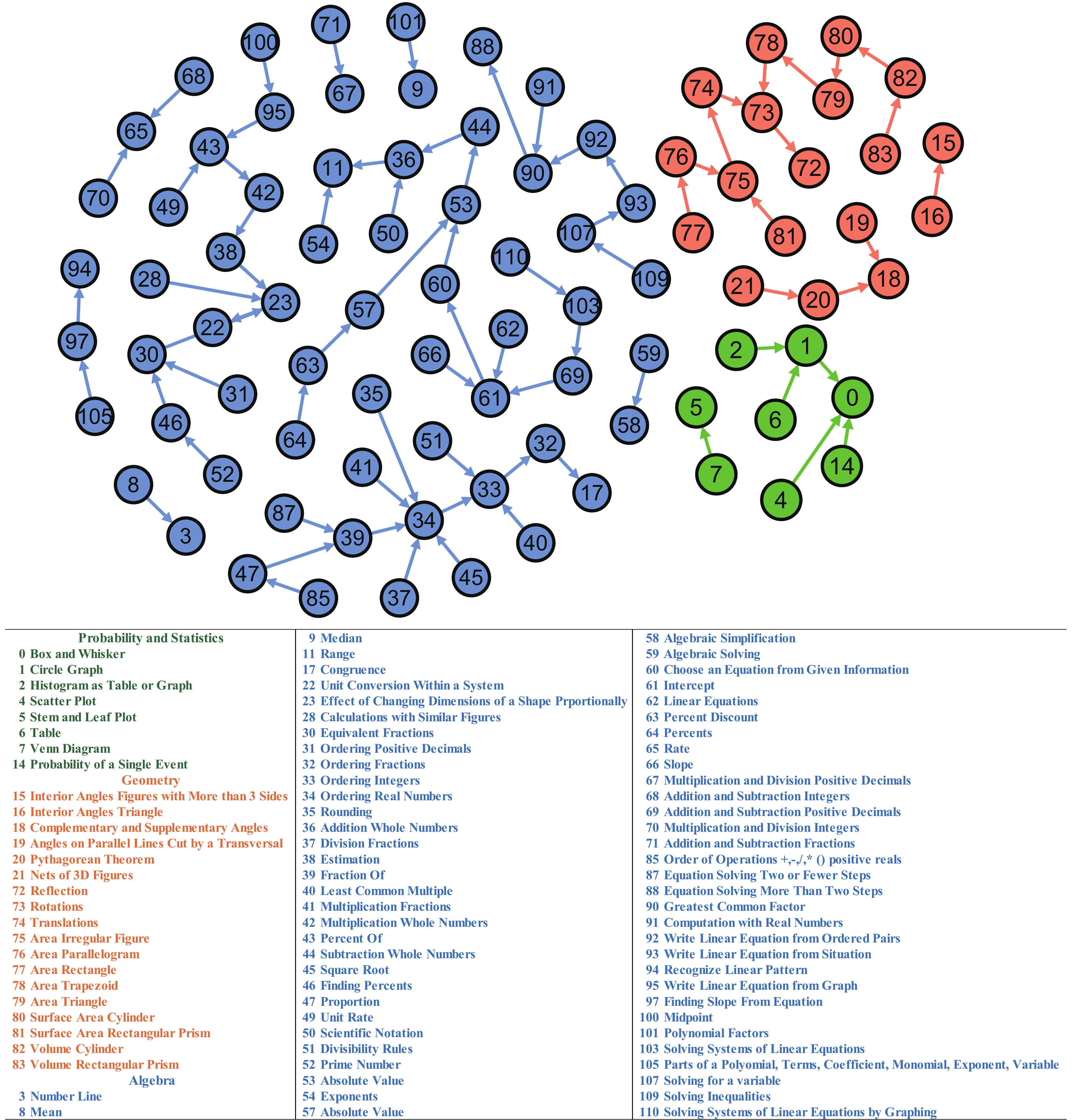}
	\caption{Visualization of concept-level relationship derived from LRP} 
	\label{fig:concept-pair}
\end{figure}
Given each question is associated with a specific math concept, we further adopt the interpreting results to explore the internal relationship among these concepts. Specifically, for each correctly predicted sequence, we have 14 relevance values on the concept associated with the last questions (i.e., the 15th question), and thus 14 directed concept pairs with a relevance value can be obtained. Note that the concept pair direction starts from the exercise concept (i.e., the exercise question) and ends at the predicted concept (i.e., the 15th question), and the pairs with two same concepts can be simply ignored. Subsequently, we collect the valid concept pairs from all the exercise sequences, and group them according to the predicted concept. We then average the absolute values of relevance for each group and accordingly find out the most relevant concept for each predicted concept. Figure~\ref{fig:concept-pair} shows the paired concept information, where three different colors are used to depict some large clusters, which can be roughly classified into algebra, geometry and statistics. Many interesting relationships can be observed from the figure. For example, \textit{node~34} (on the bottom of the blue cluster) representing concept \textit{ordering real number} is directed by a number of relevant but advanced concepts (e.g., \textit{division fraction} and \textit{square root}), while itself directs to its prerequisite concept \textit{ordering integers}. Such interesting results reflect the LRP method could, at least partially, restore the internal concept-level relationships captured by the built DLKT model, while further studies might be needed to explore their underlying meaning from the perspective of education. Note that Figure~\ref{fig:concept-pair} only depicts the identified relationships inside the clusters, and some potential relationships across the clusters are worth to be further studied as well.

\section{Conclusion}\label{sec:conclusion}
We have introduced a post-hoc interpretability method into KT domain, which is applicable to general RNN-based DLKT models. We demonstrated the promise of this approach via using its LRP method to explain a LSTM-based DLKT model, where two specific strategies are designed to compute the relevance values for weighted linear connection and multiplicative connection respectively. We conducted the experiments to validate the proposed method from the perspectives of question and concept, where the experiment results show the feasibility of using the derived relevance values for interpreting the DLKT model's predictions. On a broader canvas, we believe this work could be a solid step towards building fully interpretable and explainable DLKT models, given the fact that using deep learning for modeling learner's cognitive process might be the most promising direction for eventually solving KT problem.



%
%

\bibliographystyle{splncs04}
\bibliography{ref}

\begin{thebibliography}{10}
\providecommand{\url}[1]{\texttt{#1}}
\providecommand{\urlprefix}{URL }
\providecommand{\doi}[1]{https://doi.org/#1}

\bibitem{Saltelli04Wiley}
Andrea, S., Stefano, T., Francesca, C., Ratto, M.: Sensitivity Analysis in
  Practice:A Guide to Assessing Scientific Models. John Wiley \& Sons, Ltd
  (2004)

\bibitem{arbabzadah2016cv}
Arbabzadah, F., Montavon, G., Muller, K., Samek, W.: Identifying individual
  facial expressions by deconstructing a neural network. In: Proceedings of
  German Conference on Pattern Recognition. pp. 344--354 (2016)

\bibitem{arras2017ps}
Arras, L., Horn, F., Montavon, G., Muller, K., Samek, W.: "what is relevant in
  a text document?": An interpretable machine learning approach. Plos One
  \textbf{12}(8),  0181142--0181142 (2017)

\bibitem{arras2017explaining}
Arras, L., Montavon, G., M{\"u}ller, K.R., Samek, W.: Explaining recurrent
  neural network predictions in sentiment analysis. EMNLP 2017 p.~159 (2017)

\bibitem{bach2015po}
Bach, S., Binder, A., Montavon, G., Klauschen, F., Muller, K., Samek, W.: On
  pixel-wise explanations for non-linear classifier decisions by layer-wise
  relevance propagation. Plos One  \textbf{10}(7),  0130140 (2015)

\bibitem{baker08its}
Baker, R.S.J.d., Corbett, A.T., Aleven, V.: More accurate student modeling
  through contextual estimation of slip and guess probabilities in bayesian
  knowledge tracing. In: Proceedings of International Conference on Intelligent
  Tutoring Systems. pp. 406--415. Berlin, Germany (2008)

\bibitem{d2008more}
Baker, R.S., Corbett, A.T., Aleven, V.: More accurate student modeling through
  contextual estimation of slip and guess probabilities in bayesian knowledge
  tracing. In: International Conference on Intelligent Tutoring Systems. pp.
  406--415. Springer (2008)

\bibitem{baker2011detecting}
Baker, R.S., Goldstein, A.B., Heffernan, N.T.: Detecting learning
  moment-by-moment. International Journal of Artificial Intelligence in
  Education  \textbf{21}(1-2),  5--25 (2011)

\bibitem{cen2006learning}
Cen, H., Koedinger, K.R., Junker, B.W.: Learning factors analysis – a general
  method for cognitive model evaluation and improvement. In: Proceedings of
  International Conference on Intelligent Tutoring Systems,. pp. 164--175
  (2006)

\bibitem{chaudhry2018modeling}
Chaudhry, R., Singh, H., Dogga, P., Saini, S.K.: Modeling hint-taking behavior
  and knowledge state of students with multi-task learning. In: Proceedings of
  Educational Data Mining (2018)

\bibitem{chen2018prerequisite}
Chen, P., Lu, Y., Zheng, V.W., Pian, Y.: Prerequisite-driven deep knowledge
  tracing. In: 2018 IEEE International Conference on Data Mining (ICDM). pp.
  39--48. IEEE (2018)

\bibitem{chen2017tracking}
Chen, Y., Liu, Q., Huang, Z., Wu, L., Chen, E., Wu, R., Su, Y., Hu, G.:
  Tracking knowledge proficiency of students with educational priors. In:
  Proceedings of the 2017 ACM on Conference on Information and Knowledge
  Management. pp. 989--998. ACM (2017)

\bibitem{cho2014properties}
Cho, K., Van~Merri{\"e}nboer, B., Bahdanau, D., Bengio, Y.: On the properties
  of neural machine translation: Encoder-decoder approaches. arXiv preprint
  arXiv:1409.1259  (2014)

\bibitem{corbett1994knowledge}
Corbett, A.T., Anderson, J.R.: Knowledge tracing: Modeling the acquisition of
  procedural knowledge. User modeling and user-adapted interaction
  \textbf{4}(4),  253--278 (1994)

\bibitem{Halpern18EDM}
David, H., et~al.: knowledge tracing using the brain. In: Proceedings of the
  Educational Data Mining (EDM) (2018)

\bibitem{deng2019IJDSA}
Deng, H.: Interpreting tree ensembles with intrees. International Journal of
  Data Science and Analytics  \textbf{7},  277--287 (2019)

\bibitem{ding2017acl}
Ding, Y., Liu, Y., Luan, H., Sun, M.: Visualizing and understanding neural
  machine translation. In: Proceedings of the 55th Annual Meeting of the
  Association for Computational Linguistics. vol.~1, pp. 1150--1159 (2017)

\bibitem{du2018towards}
Du, M., Liu, N., Song, Q., Hu, X.: Towards explanation of dnn-based prediction
  with guided feature inversion. In: Proceedings of the 24th ACM SIGKDD
  International Conference on Knowledge Discovery \& Data Mining. pp.
  1358--1367 (2018)

\bibitem{feng2009addressing}
Feng, M., Heffernan, N., Koedinger, K.: Addressing the assessment challenge
  with an online system that tutors as it assesses. User Modeling and
  User-Adapted Interaction  \textbf{19}(3),  243--266 (2009)

\bibitem{monta18DSP}
Grégoire, M., Wojciech, S., Klaus-Robert, M.: Methods for interpreting and
  understanding deep neural networks. Digital Signal Processing  \textbf{73},
  1--15 (2018)

\bibitem{hochreiter1997long}
Hochreiter, S., Schmidhuber, J.: Long short-term memory. Neural Computation
  \textbf{9}(8),  1735--1780 (1997)

\bibitem{lapuschkin2016cvpr}
Lapuschkin, S., Binder, A., Montavon, G., Muller, K., Samek, W.: Analyzing
  classifiers: Fisher vectors and deep neural networks. In: Proceedings of IEEE
  Conference on Computer Vision and Pattern Recognition (CVPR). pp. 2912--2920
  (2016)

\bibitem{liu2018kdd}
Liu, Q., Huang, Z., Huang, Z., Liu, C., Chen, E., Su, Y., Hu, G.: Finding
  similar exercises in online education systems. In: Proceedings of
  International Conference on Knowledge Discovery and Data Mining. pp.
  1821--1830 (2018)

\bibitem{liu2017towards}
Liu, R., Koedinger, K.R.: Towards reliable and valid measurement of
  individualized student parameters. In: Proceedings of the 10th International
  Conference on Educational Data Mining. pp. 135--142 (2017)

\bibitem{melis2018towards}
Melis, D.A., Jaakkola, T.S.: Towards robust interpretability with
  self-explaining neural networks. In: Proceedings of Advances in Neural
  Information Processing Systems(NIPS). pp. 7786--7795 (2018)

\bibitem{pardos2013adapting}
Pardos, Z.A., Bergner, Y., Seaton, D.T., Pritchard, D.E.: Adapting bayesian
  knowledge tracing to a massive open online course in edx. EDM  \textbf{13},
  137--144 (2013)

\bibitem{pardos2011kt}
Pardos, Z.A., Heffernan, N.T.: Kt-idem: introducing item difficulty to the
  knowledge tracing model. In: International Conference on User Modeling,
  Adaptation, and Personalization. pp. 243--254. Springer (2011)

\bibitem{pavlik2009performance}
Pavlik~Jr, P.I., Cen, H., Koedinger, K.R.: Performance factors analysis--a new
  alternative to knowledge tracing. In: Proceedings of International Conference
  on Artificial Intelligence in education (2009)

\bibitem{piech2015deep}
Piech, C., Bassen, J., Huang, J., Ganguli, S., Sahami, M., Guibas, L.J.,
  Sohl-Dickstein, J.: Deep knowledge tracing. In: Advances in Neural
  Information Processing Systems. pp. 505--513 (2015)

\bibitem{poulin2006visual}
Poulin, B., Eisner, R., Szafron, D., Lu, P., Greiner, R., Wishart, D.S., Fyshe,
  A., Pearcy, B., Macdonell, C., Anvik, J.: Visual explanation of evidence in
  additive classifiers. In: Proceedings of National Conference on Artificial
  Intelligence. pp. 1822--1829 (2006)

\bibitem{qiu2011edm}
Qiu, Y., Qi, Y., Lu, H., Pardos, Z.A., Heffernan, N.T.: Does time matter?
  modeling the effect of time with bayesian knowledge tracing. In: Proceedings
  of Educational Data Mining Workshop at the 11th International Conference on
  User Modeling. pp. 139--148 (2011)

\bibitem{rumelhart1988learning}
Rumelhart, D.E., Hinton, G.E., Williams, R.J.: Learning representations by
  back-propagating errors. Nature  \textbf{323}(6088),  696--699 (1988)

\bibitem{santoro2016icml}
Santoro, A., Bartunov, S., Botvinick, M., Wierstra, D., Lillicrap, T.:
  Meta-learning with memory-augmented neural networks. In: Proceedings of
  International Conference on Machine Learning. pp. 1842--1850 (2016)

\bibitem{schuster1997bidirectional}
Schuster, M., Paliwal, K.K.: Bidirectional recurrent neural networks. IEEE
  transactions on Signal Processing  \textbf{45}(11),  2673--2681 (1997)

\bibitem{strumbelj2010an}
Strumbelj, E., Kononenko, I.: An efficient explanation of individual
  classifications using game theory. Journal of Machine Learning Research
  \textbf{11},  1--18 (2010)

\bibitem{su2018exercise}
Su, Y., Liu, Q., Liu, Q., Huang, Z., Yin, Y., Chen, E., Ding, C., Wei, S., Hu,
  G.: Exercise-enhanced sequential modeling for student performance prediction.
  In: Thirty-Second AAAI Conference on Artificial Intelligence (2018)

\bibitem{thai2012factorization}
Thai-Nghe, N., Drumond, L., Horv{\'a}th, T., Krohn-Grimberghe, A., Nanopoulos,
  A., Schmidt-Thieme, L.: Factorization techniques for predicting student
  performance. In: Educational recommender systems and technologies: Practices
  and challenges, pp. 129--153. IGI Global (2012)

\bibitem{vie2019knowledge}
Vie, J., Kashima, H.: Knowledge tracing machines: Factorization machines for
  knowledge tracing. In: Proceedings of AAAI Conference on Artificial
  Intelligence. vol.~33, pp. 750--757 (2019)

\bibitem{wang2016structured}
Wang, Z., Zhu, J., Li, X., Hu, Z., Zhang, M.: Structured knowledge tracing
  models for student assessment on coursera. In: Proceedings of the Third
  (2016) ACM Conference on Learning@ Scale. pp. 209--212 (2016)

\bibitem{xiong2016going}
Xiong, X., Zhao, S., Van~Inwegen, E., Beck, J.: Going deeper with deep
  knowledge tracing. In: EDM. pp. 545--550 (2016)

\bibitem{yang2018implicit}
Yang, H., Cheung, L.P.: Implicit heterogeneous features embedding in deep
  knowledge tracing. Cognitive Computation  \textbf{10}(1),  3--14 (2018)

\bibitem{yazdani1989air}
Yazdani, M.: Intelligent tutoring systems survey. Artificial Intelligence
  Review  \textbf{1}(1),  43--52 (1989)

\bibitem{yeung2018addressing}
Yeung, C.K., Yeung, D.Y.: Addressing two problems in deep knowledge tracing via
  prediction-consistent regularization. In: Proceedings of the Fifth Annual ACM
  Conference on Learning at Scale. p.~5. ACM (2018)

\bibitem{yeung2019edm}
Yeung, C.: Deep-irt: Make deep learning based knowledge tracing explainable
  using item response theory. In: Proceedings of Educational Data Mining (2019)

\bibitem{yudelson2013individualized}
Yudelson, M.V., Koedinger, K.R., Gordon, G.J.: Individualized bayesian
  knowledge tracing models. In: International Conference on Artificial
  Intelligence in Education. pp. 171--180. Springer (2013)

\bibitem{zeiler2014eccv}
Zeiler, M.D., Fergus, R.: Visualizing and understanding convolutional networks.
  In: Proceedings of European Conference on Computer Vision. pp. 818--833
  (2014)

\bibitem{zhang2017dynamic}
Zhang, J., Shi, X., King, I., Yeung, D.Y.: Dynamic key-value memory networks
  for knowledge tracing. In: Proceedings of the 26th International Conference
  on World Wide Web. pp. 765--774 (2017)

\end{thebibliography}
\end{document}